\documentclass{article}
\usepackage{amsmath}
\usepackage{graphicx}
\usepackage{comment}
\usepackage{subfig}
\usepackage{xfrac}
\usepackage{amssymb}
\usepackage[final]{corl_2019} 

\title{Multimodal Attention Branch Network for Perspective-Free Sentence Generation}

\author{
  Aly Magassouba \And Komei Sugiura \And  Hisashi Kawai\\
  National Institute of Information and Communications Technology\\ 
  Japan\\
  \texttt{name.surname@nict.go.jp} \\
}

\begin{document}
\maketitle
\graphicspath{{figures/}}



\begin{abstract}
In this paper, we address the automatic sentence generation of fetching instructions for domestic service robots. Typical fetching commands  such as ``bring me the yellow toy from the upper part of the white shelf'' includes referring expressions, {\it i.e.}, ``from the white upper part of the white shelf''. To solve this task, we propose a multimodal attention branch network (Multi-ABN) which generates natural sentences in an end-to-end manner. Multi-ABN uses multiple images of the same fixed scene to generate sentences that are not tied to a particular viewpoint. This approach combines a linguistic attention branch mechanism with several attention branch mechanisms. We evaluated our approach, which outperforms the state-of-the-art method on a standard metrics.  Our method also allows us to visualize the alignment between the linguistic input and the visual features. 
\end{abstract}
\section{Introduction}
The growth in the aged population has steadily increased the need for daily care and support.  Robots that can physically assist people with disabilities \citep{brose2010role} offer an alternative to overcoming the shortage of home care workers. This context has boosted the need for standardized domestic service robots (DSRs) that can provide necessary support functions  as shown by \citep{piyathilaka2015human, smarr2014domestic,iocchi2015robocup}. 

Nonetheless, one of the main limitations of DSRs is their inability to naturally interact through language. Specifically, most DSRs do not allow users to instruct them with various expressions relating to an object for fetching tasks. By tackling this limitation, a user-friendly way to interact with DSRs could be provided to non-expert users.

Solving this task is particularly important for robots that perform manipulation tasks in home environments. Indeed, to understand ambiguous carry-and-place \citep{magassouba2018multimodal} or fetching \citep{magassouba2019understanding}  instructions in an end-to-end approach, a large number of samples of natural manipulation instructions are required. Unfortunately, it is costly to obtain such data. Hence, methods to automatically augment or generate instructions data could drastically reduce the cost of building a large-scale dataset for DSRs.

In this light, our work addresses the task of automatic sentence generation for fetching instructions. This task consists of generating various natural fetching instructions given a target object in a image, {\it e.g.},``{\it Go get me the empty bottle from the armchair on the right side }.'' Natural sentences  often contain referring expressions to designate a given target. However, generating referring expressions is challenging. Indeed, the many-to-many nature of mapping between the language and real world makes it difficult to generate such sentences but, at the same time, offers flexible sentence structure.

In this paper, we propose the multimodal attention branch network (Multi-ABN) which is an extension of the attention branch network proposed in \citep{Fukui_2019_CVPR}. The initial attention branch network was proposed as an image classifier, inspired by class activation mapping (CAM) \citep{zhou2016cvpr} structures, to infer attention maps. It is composed of an attention branch that predicts an attention map and a perception branch that classifies images. This architecture is extended in Multi-ABN where several visual and linguistic attention branches are proposed to respectively infer the visual and linguistic attention maps.  Indeed, instead of using a single image of a given scene, several snapshots of the same scene from different viewpoints are processed to generate perspective-free referring expressions. Our aim is to generate sentences that are not tied to the viewpoint of the human-annotated training image. From these attention maps, a long short-term memory (LSTM) network generates fetching instructions in the perception branch.

The main contributions of this paper are summarized as:
\begin{itemize}
\item[$\bullet$] We propose a Multi-ABN which generates fetching instructions based on multiple images from different perspectives of a fixed scene  \ref{sec:method}. 
\item[$\bullet$] Multi-ABN extends the existing methods by adopting visual and linguistic attention mechanisms based on class activation mapping structures.
\item[$\bullet$] Multi-ABN outputs a visual explanation for the generated fetching instructions. 
\end{itemize}	
\section{Related work}
Building communicative robots that can understand ambiguous manipulation instructions generally require the fusion of multiple modalities, which are generally visual and linguistic. Several studies focus on understanding manipulation instructions in an end-to-end approach. For instance, \citep{hatori2018interactively} proposed a target object prediction method from natural language in a pick-and-place task environment, using a visual semantic embedding model. Similarly \citep{Shridhar-RSS-18} tackled the same kind of problem using a two-stage model to predict the likely target from the language expression and the pairwise relationships between different target candidates. More recently, in a context related to DSRs, \citep{magassouba2019understanding} proposed to use both the target and source candidates to predict the likely target in a supervised manner. In \citep{magassouba2019understanding}, the placing task was addressed through a GAN classifier network predicting the most likely destination from the initial instruction.

The proposed systems mainly focus on multimodal language grounding through referring expression comprehension. Complementary to  these works, some recent studies have also focused on generating referring expressions to identify a target. In \citep{kunze2017spatial}, the authors proposed several algorithms to generate referring expressions in a rule-based approach. In contrast, in \citep{dougan2019learning}, the authors used deep learning for estimating spatial relations to describe an object in a sentence. However, the set of spatial relationships is hand-crafted and known beforehand. We, instead, target an end-to-end approach that do not require hand-crafted or rule-based methods.



Multi-ABN is inspired by the attention branch network (ABN) \citep{Fukui_2019_CVPR}. The ABN is based on the CAM structure \citep{zhou2016cvpr, Selvaraju_2017_ICCV} to build visual attention maps for image classification. In essence, a CAM is built to identify salient regions used by a given class in an image classifier. Attention mechanisms have also been used in different ways in image processing and natural language processing. In the context of image captioning, the authors of \citep{xu2015show} proposed to generate image captions with hard and soft visual attention. This approach learns the alignment between the salient area of an image and the generated sequence of words. Multiple visual attention networks have also been proposed in \citep{Yang_2016_CVPR} to solve visual question answering. However, most of these approaches use only a single modality for attention: visual attention. In contrast, we claim in this work that both linguistic and visual branch attention improve the sentence generation process.

To do so, we use annotated data obtained from the simulation environment SIGVerse \cite{inamura2013development}. Nowadays, many studies use simulated environments to collect synthetic data. Synthetic data tend to be increasingly photo-realistic and have the advantage  of task repeatability  as well as  environment variation for a relatively low cost. Using such environments, various tasks such as grasping \cite{bousmalis2018using} or motion control \cite{tan2018sim} have been addressed.

\section{Problem statement}
 \label{sec:statement}
\subsection{Task Description}
This study targets the generation of sentences for fetching instructions including referring expressions. Referring expressions usually describe an object using properties of the object with respect to landmark objects. A typical generated fetching instruction can be to `` go get me the pink doll on the upper part of the shelf''. In this instruction, the landmark is ``the shelf'' while ``upper part'' is a spatial referring expression. 
To generate such a sentence, our system assumes the following inputs and outputs:
\begin{itemize}
      \item[$\bullet$]{\bf Input}: a fixed scene observed from several perspectives.
      \item[$\bullet$]{\bf Output}: the most likely generated sentence for a given target and source 
\end{itemize}
The inputs of our system are more thoroughly described in Section \ref{sec:method}. The terms {\it target} and {\it source} are defined as follows.
 \begin{itemize}
     \item[$\bullet$]{\bf Target}: the daily life object ({\it e.g.}, apple or bottle) that the user intends for the robot to fetch.
     \item[$\bullet$]{\bf Source}: the origin of the target, generally pieces of furniture such as shelves or drawers.
\end{itemize}
Unlike most methods proposed in the literature, our sentence generation method is based on several images of a fixed scene. Indeed, using single image to build a fetching instruction introduces a drawback  when considering DSRs for manipulation task. This limitation is mainly related to DSRs that interact in a three-dimensional environment. There might be a mismatch between the user's perspective (a single 2D image) and the robot's perspective \cite{cohen2019grounding}. A DSR's view of a given scene is dynamic, {\it e.g.}, a target can be behind, on the left side, or on right side of the same landmark depending on the current pose.  
Hence, to avoid generating referring expressions that are related to a given point of view of the scene, {\it e.g}, ``the apple on the left side of the table'', we use images from different perspectives of the same fixed scene. In this configuration, referring expressions such as ``left of'' or ``right side of'' are correct only if they are valid for all observations. 

Several challenges should be tackled to generate valid fetching instructions. First, several objects may be of the same type as that of the target, so referring expressions should be used to disambiguate the target from the other objects. Second, several existing objects and sources  may be used as landmarks for generating the referring expressions.  However, the generated sentence should use referring expressions that do not imply any ambiguity of the target, independently of the point of view. 

The standard evaluation metrics of our approach are based on the  automatic metrics of image captioning that is BLEU, ROUGE, CIDEr and METEOR, as reported in the experimental section.

\begin{figure}[tp]
  \centering
   \subfloat{\includegraphics[scale = 0.14]{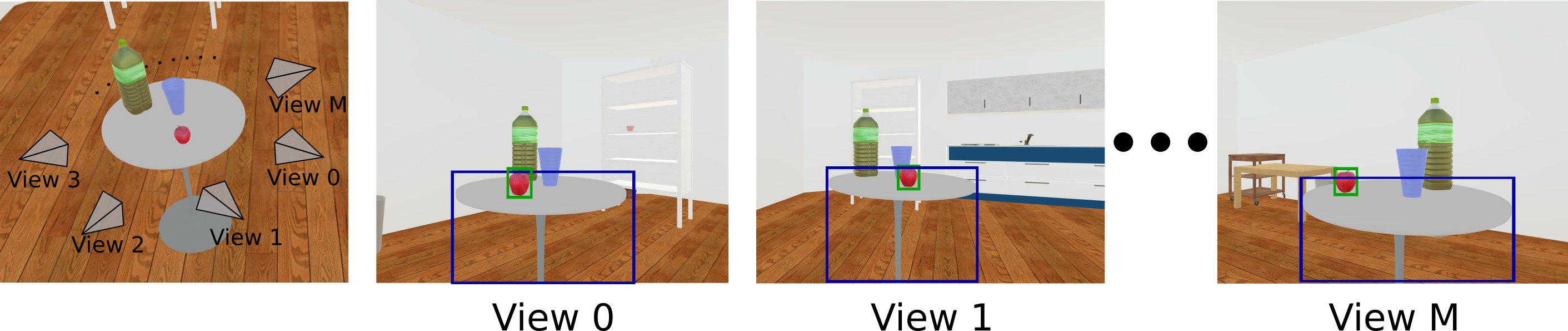}}
    \caption{\small Source (blue) and target (green) samples of the WRS-VS dataset considering several perspectives. The perspective influences the validity of the instruction and the referring expressions that can be used.  {\bf Valid sentence} : ``Bring me the apple that is near the glass on the kitchen table''/ {\bf Invalid sentence}: `` Bring me the apple on the left side of the blue glass''.} \label{fig:samples} 
\end{figure}

\subsection{Task Environments}
The sentence generation system should be general  and flexible enough to be used for various scenarios. We therefore consider a simulated environment in which the task repeatability and various situations can be addressed at low cost. In this study, we use the simulated environments that were provided in the World Robot Summit 2018 Virtual Space (WRS-VS) challenge. The simulator is based on SIGVerse \cite{inamura2013development}, which is a three-dimensional environment based on the Unity engine and is able to simulate interactions between agents and the environment. The WRS-VS consists of typical indoor environments as illustrated in Fig. \ref{fig:samples}, from which we built a dataset. 
In this environment, we use a DSR that records several snapshots of a given observable scene. 
From this context, our method should generate fetching instructions such as ``{\it Give me the rabbit doll from the upper part of the shelf}''. 
\begin{figure*}[t]
   \centering
      \includegraphics[scale=0.152]{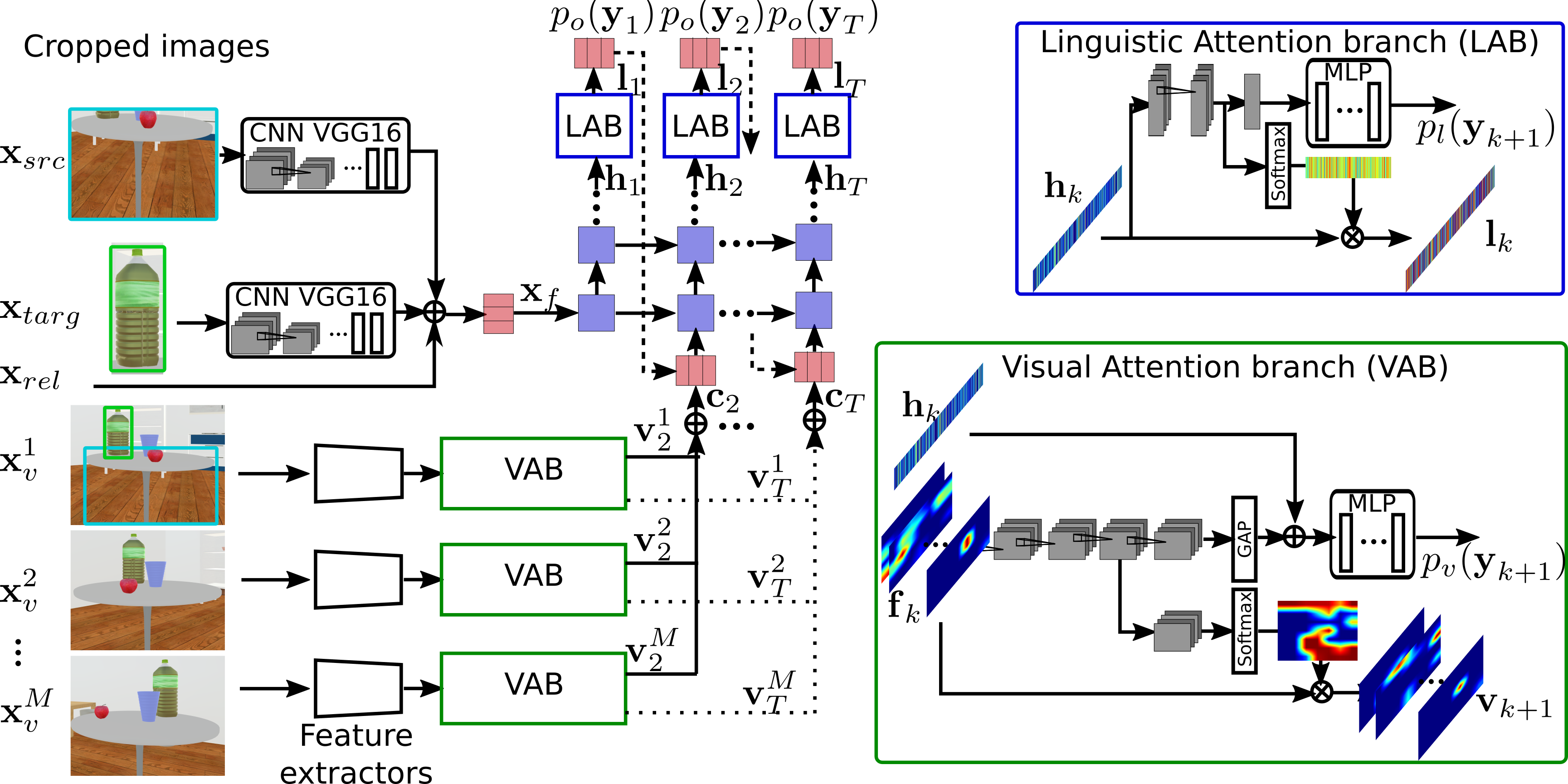}
      \caption{Proposed method framework: Multimodal attention branch mechanism  that couples a linguistic attention branch to visual attention branches to generate fetching instructions}
   \label{fig:abn}
 \end{figure*}

\section{Proposed method}
Multi-ABN is composed of a linguistic attention branch as well as several visual attention branches for each different viewpoint. In the following we detail the input features, as well as the different branches (attention and perception branches) that have been used to address multimodality. The full network structure is given in Fig. \ref{fig:abn}. The aim of this network is to generate a sequence $Y = \{ {\bf y}_1, {\bf y}_2 \dots {\bf y}_T \}$  where $T$ is the length of a generated sentence and ${\bf y}_k \in \mathbf{R}^{d}$ for an embedding dimension $d$.
\label{sec:method}
 \subsection{Input features}
Let us consider $X = \{ {\bf x}_n | n = 1, \dots, N \}$ a dataset composed of $N$ samples.  Hereinafter, for readability, we voluntarily omit the sample index $n$, so that ${\bf x}_n$ is written as ${\bf x}$  when further clarity is not required. Each sample ${\bf x}$ is characterized by the set of inputs:
\begin{equation}\label{equ:inputs}
    {\bf x}=\{ {\bf x}_{v}^{1}, {\bf x}_{v}^{2}, \dots, {\bf x}_{v}^{M}, {\bf x}_{src}, {\bf x}_{targ}, {\bf x}_{rel}\}.
\end{equation}

The input ${\bf x}_{v}^{j}$ defines the set of image inputs taken from different viewpoints, where  the superscript $j \in \{1, \dots,  M \}$ defines the camera ID. Additionally,  ${\bf x}_{src}$, ${\bf x}_{targ}$ and ${\bf x}_{rel}$ respectively denote the source image, target image and relation features of the target in the environment. It should be noted that the source, target and relation features are extracted once only from the main image ${\bf x}_{v}^{1}$.

Inputs ${\bf x}_{targ}$ and ${\bf x}_{src}$ are the cropped images of the target and source respectively. Inputs ${\bf x}_{rel}$  denotes the target-source, target-image and source-image spatial relation.  Each of these relations are characterized by the following: 
\begin{equation}\label{equ:x}
{\bf r}_{\sfrac{m}{n}}=\begin{bmatrix}
\frac{x_m}{W_n}, & \frac{y_m}{H_n}, &\frac{w_m}{W_n},&\frac{h_m}{H_n},&\frac{w_m h_n}{W_m H_n}
\end{bmatrix}
\end{equation}
where $(x_m, y_m, w_m, h_m)$ are the horizontal position, vertical position, width and height of the component $m$, while $W_n$ and $H_n$ are the width and height of the component $n$. As a result, the relation features are defined as ${\bf x}_{rel} = \{{\bf r}_{\sfrac{targ}{src}}, {\bf r}_{\sfrac{targ}{v}}, {\bf r}_{\sfrac{src}{v}}\}$ with a dimension $d_{rel}=15$.

\subsection{Attention branches}
\subsubsection{Visual attention branches}
The attention branch \citep{Fukui_2019_CVPR} allows us to build both the linguistic and visual attention map, based on the extracted feature maps. We consider two different types of feature maps, linguistic and visual detailed below.

Multi-ABN is composed of $M$ visual attention branches, that is for the images $ {\bf x}_{v}^{j}$. Each of these visual attention branch takes as input visual feature maps denoted ${\bf f}_k$. These feature maps are obtained from a convolutional feature extraction of $ {\bf x}_{v}^{j}$ . In this paper, we based our feature extractor on VGG16\citep{simonyan2014very}. Note that other feature extractor such as ResNet\citep{he2016deep} could also be used. ${\bf f}_k$ then corresponds to the output from the $5^{th}$ convolutional block of the VGG network. Each feature map ${\bf f}_k$ has a dimension $14\text{x}14\text{x}512$. 
A visual attention branch outputs visual feature maps ${\bf v}_{k+1}$ weighted by a visual attention mask.
To do so, inspired by the CAM structure \citep{zhou2016cvpr}, the visual feature maps are encoded through four convolutional layers. These convolutional layers are followed by a global average pooling (GAP) and a two-layer multilayer perceptron (MLP) denoted as $\text{MLP}_a$. 
Prior to the first layer of $MLP_a$, the visual features are concatenated with the linguistic feature map ${\bf h}_k$ (see next section). The likelihood $p_v({\bf y}_{k+1})$ is then predicted.
In parallel, a visual attention map ${\bf a}_k$ is created by an additional convolution and sigmoid normalization of the third convolutional layer of the visual attention branch.  This attention map  allows to selectively focus on certain parts of an image related to the predicted sequence. The output visual feature maps are then obtained by a masking process given by:
\begin{equation}\label{equ:vis_att}
    {\bf v}_{k}= {\bf a}_k \odot {\bf f}_{k},
\end{equation}
where $\odot$ denotes the Hadamard product.
\subsubsection{Linguistic attention branch}
In addition a  linguistic attention branch takes as input the linguistic feature maps ${\bf h}_{k}$.
 ${\bf h}_{k}$ is simply defined as the output (or hidden state) of an LSTM generating the instruction sequence $Y$. This LSTM network is detailed in the next section. The linguistic feature map ${\bf h}_{k}$ is encoded through 1-dimensional convolution layers followed by a single fully connected layer so as to output likelihood $p_l({\bf y}_{k+1})$. Linguistic attention map ${\bf a}_l$ is obtained from the second convolutional layer  that is convoluted in an additional layer and normalized by a sigmoid activation function. This attention map allows to selectively focus on a area of the LSTM state the also encodes all the previous states. 
 Similarly to visual attention branches, the output ${\bf l}_{k}$ of linguistic attention branch is given by:
 \begin{equation}\label{equ:ling_att}
    {\bf l}_{k}= {\bf a}_l \odot {\bf h}_{k}
\end{equation}

\subsection{Perception branch}
The perception branch is a classifier that predicts the likelihood of $p({\bf y}_{k+1})$ in a sequence of length $T$. The perception branch takes as input the concatenation  of all weighted visual feature maps ${\bf v}_{k}^{j}$  that is referred as ${\bf c}_{k}$ and the weighted linguistic feature map ${\bf l}_k$, as well as the target, source and relation features. The architecture of the perception branch is based on a multilayer LSTM network. The perception also outputs linguistic feature map ${\bf h}_{k}$  that is simply the hidden state of each LSTM cell. Note that because each embedded word ${\bf y}_k$ is predicted sequentially, the last hidden state also corresponds to the output of the LSTM. More thoroughly, the LSTM is initialized by the latent space feature ${\bf x}_{f}$ obtained by embedding and concatenating  the target ${\bf x}_{targ}$, the source ${\bf x}_{src}$ and the relation feature ${\bf x}_{rel}$. In a compact formulation, the first hidden state can be written as 
\begin{equation}\label{equ:LSTM_init}
    {\bf h}_1= \text{LSTM}({\bf x}_{f}).
\end{equation}
It should be  mentioned that the forget, memory, output and hidden state variables are voluntarily omitted for more concision.  In the following steps, considering an iteration $k$, with $k>0$, each hidden state is defined as follows
\begin{equation}\label{equ:LSTM_cells}
    {\bf h}_k=\text{LSTM}(E({\bf c}_{k} \oplus {\bf y}_{k-1})),
\end{equation}
where $\oplus$ indicates a concatenation operation and $E(\cdot)$ is an embedding function. In this configuration, ${\bf c}_{k}$ can be considered as the visual context of the current LSTM state. Eventually, to predict the likelihood $p_o({\bf y}_{k+1})$ in the sequence, the weighted linguistic feature map ${\bf l}_{k}$ is processed in a embedding layer.

\subsection{Loss functions}
The global training loss function of the network is the sum of the attention branch loss $L_{att}$ and the perception branch loss $L_{per}$ so that
\begin{equation}\label{equ:loss}
    L = L_{att} + L_{per}.
\end{equation}
Perception loss $L_{per}$ is defined as a cross-entropy  function in which the class of ${\bf y}_{k+1}$ is predicted through
\begin{align} \label{equ:J}
    L_{per} &= -\sum_n \sum_{m} y^{*}_{nm} \log p(y_{nm}),
\end{align}
where $y^{*}_{nm}$ denotes the label given to the $m$-th dimension of the $n$-th sample, and $y_{nm}$ denotes its prediction.
The attention loss $L_{att}$ depends on the visual attention loss and linguistic attention loss, which are also both cross-entropy loss functions, as defined in Eq. \eqref{equ:J}, that enable to build the corresponding attention maps.

\section{Experiments}
\label{sec:experiment}
\subsection{Dataset}
We evaluated our method with the WRS-VS dataset introduced in Section \ref{sec:statement} and illustrated in Fig. \ref{fig:samples}. We collected $308 \times \text{M}$ images. In the following experiment, we set $M = 3$, which means that there were three different images from each given scene. We annotated $1015$ targets with $2015$ sentences in the training set and $34$ targets with $74$ different sentences in the validation set.
The annotation was performed by an expert user. This data set has an average of $3.4$ targets per image, and $9.5$ words for each instruction. The vocabulary set $V$ is composed of 233 unique words.

\subsection{Experimental Setup}
The parameter settings of the Multi-ABN are summarized in Table \ref{tab:param}.
We describe first the different attention branches that compose the Multi-ABN.
Each visual attention branch (noted Vis. AB) uses 2D convolutional layers of size $3 \times 3 \times \|V\|$ before the global average pooling layer. To generate each visual attention map, a convolutional layer of dimension $1 \times 1 \times 1$ is used.
Because we consider $M = 3$ different images, a two-layer $MLP_{a}$ is used to encode the different weighted visual feature maps concatenated with the linguistic feature map. 
In parallel, the linguistic attention branch (noted Ling. AB) uses 1D convolutional layers of size $3\times 3\times \|V\|$ followed by a single-layer embedding of dimension $\|V\|$. Similarly to the visual case, the linguistic attention map is obtained by processing the features with a convolutional layer of dimension $1\times 1\times 1$.

\begin{table}[h]
\small
\centering
\caption{\small Parameter settings of the  Multi-ABN}\label{tab:param}
\begin{tabular}{|c|l|}
\hline
Multi-ABN Opt. method & Adam (Learning rate= $5e^{-4}$, $\beta_1=0.99$, $\beta_2=0.9$) \\
\hline
LSTM   &$3$ layers, $1024$-cell\\
\hline
MLP num. nodes &  $MLP_{a}$: $1024$, $1024$ \\
\hline
Vis. AB & Conv: $3\text{x}3\text{x}\|V\|$, Att. Conv : $1\text{x}1\text{x}1$\\
\hline
Ling.  AB & Conv: $3\text{x}3\text{x}\|V\|$, Att. Conv : $1\text{x}1\text{x}1$, MLP: $\|V\|$ \\ 
\hline
Batch size&$32$ \\
\hline
\end{tabular}
\end{table}

In the perception branch, the LSTM has $N =3$ layers, with each cell having dimension $d=1,024$. The network is trained with an Adam optimizer with a learning rate of $5e^{-4}$, considering a batch size of 32 samples.

\subsection{Quantitative results}
As mentioned in Section \ref{sec:statement}, we use standard image captioning metrics to evaluate the performance of Multi-ABN. To be exhaustive, we report the results of baseline metrics BLEU score (1-gram to 4-gram) as well as more evolve scoring systems ROUGE, CIDEr and METEOR. These scores are reported in each column of Table \ref{tab:results}. 

The Multi-ABN was compared with a baseline method \citep{yu2017joint}, in which a speaker model is used to generate sentences. For fair comparison, because the speaker model is only adapted to $M = 1$ images, we concatenated $M = 3$ images into a single one  to form the input of the speaker model. The same method was applied for comparison to another baseline that is the visual semantic embedding architecture \cite{vinyals2015show}.
As reported in Table \ref{tab:results}, Multi-ABN outperforms the speaker model under ROUGE, CIDEr and BLEU score, while METEOR evaluation does not make emphasize any significant difference. In comparison to VSE method, Multi-ABN performs significantly better for all metrics, in particular, CIDEr score is improved by 0.46 points.

\begin{table}[t]
\normalsize
\caption{\small Evaluation of Multi-ABN sentence generation. The Multi-ABN is compared with  speaker model\cite{yu2017joint} using reinforcement learning as well as a baseline method using visual semantic embedding (VSE)\cite{vinyals2015show},  Multi-ABN with visual attention branch (VAB) only, and Multi-ABN with linguistic attention branch (LAB) only.}
\label{tab:results}
\centering
\begin{tabular}{l|ccccccc}
\hline
{\bf Method }& \multicolumn{7}{c}{Evaluation metric} \\
\cline{2-8}
 &\multicolumn{1}{c|}{BLEU-1} &\multicolumn{1}{c|}{BLEU-2} &\multicolumn{1}{c|}{BLEU-3} &\multicolumn{1}{c|}{BLEU-4} &\multicolumn{1}{c|}{ROUGE} &\multicolumn{1}{c|}{METEOR} &\multicolumn{1}{c}{CIDEr} \\
\hline
\hline
Speaker \cite{yu2017joint}& 0.319 & 0.201 &0.132 & 0.102 & 0.309 & {\bf 0.195} &0.802   \\
\hline
VSE & 0.306&	0.199&	0.123&	0.073&	0.285& 0.108 & 0.588  \\
\hline
Ours (VAB only) & 0.323 & 0.216 & 0.143 & 0.102 & 0.333 & 0.165 & 0.824  \\
\hline
Ours (LAB only) & 0.301 &	0.250 &	0.123 &	0.099 &	0.353 & 0.142 & 0.902 \\
\hline
Ours (Multi-ABN)  & \bf 0.390 &	\bf 0.287&	\bf 0.184&	\bf 0.142&	\bf 0.359& 0.193 & \bf 1.048  \\
\hline
\end{tabular}
\end{table}

In addition, several ablation tests were conducted to isolate and emphasize the contribution of each attention branch mechanism. We compare our results, Multi-ABN with visual attention branch only (VAB) and Multi-ABN with linguistic attention branch (LAB).
The results in Table \ref{tab:results} show that the Multi-ABN drastically improved all the metrics. The visual and linguistic attention branches each improved the baseline visual semantic architecture. The LAB improved the sentence generation quality more in terms of ROUGE and CIDEr, while the BLEU and METEOR scores were better with the VAB only. This suggests that LAB leads to a better generalization and variety in the produced sentence, while the VAB is better able to mimic the dataset sentences.

\subsection{Qualitative results} 

\subsubsection{Sentence generation}
\begin{figure*}[h]

 \subfloat[]{\includegraphics[height=2.4cm, width=3.3cm]{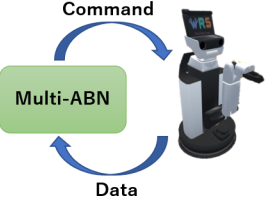}}
\enskip
 \subfloat[I want a rabbit item on the upper part of shelf]{\includegraphics[scale = 0.19]{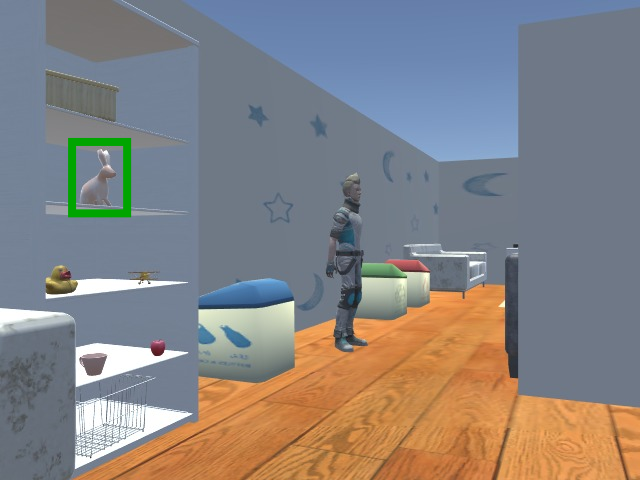}}
\enskip
 \subfloat[I want an empty large plastic bottle placed on the left hand sofa]{\includegraphics[scale = 0.19]{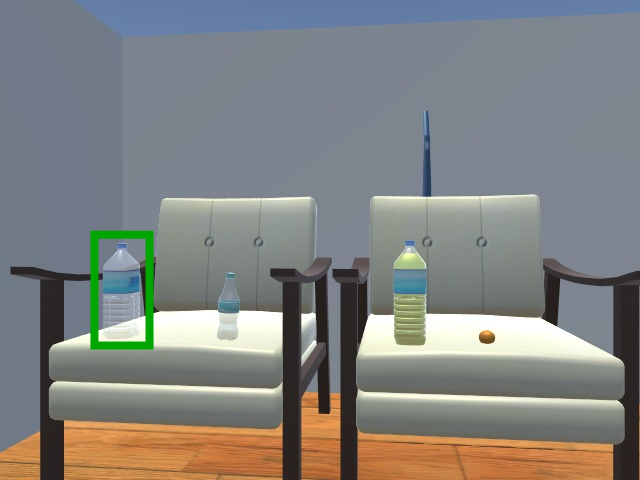}} \enskip
 \subfloat[Can you get a large plastic bottle on the side table]{\includegraphics[scale = 0.19]{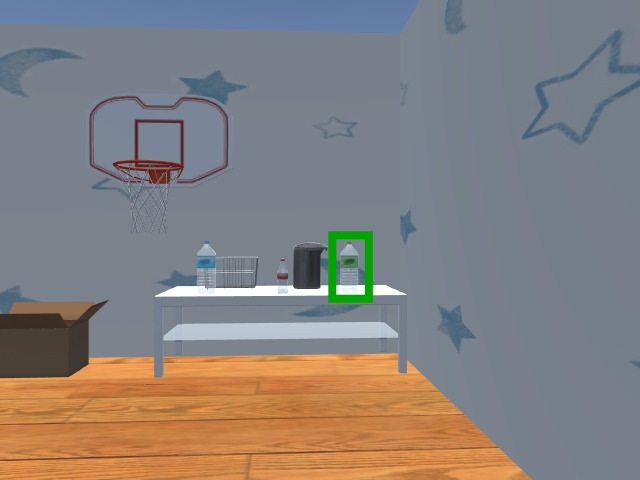}}
 \caption{\small Generated sentences by our method. Solid green rectangles represent the target. Only the first image ${\bf x}_v^{1}$ of each sample are given. Subfigures (b) and (c) show correct predictions while subfigure (d) shows an erroneous generation}
    \label{fig:wrs_samples}
\end{figure*}

In the following, because of limited space, we illustrate our results with only a single image ({\it i.e.}, ${\bf x}_v^{1}$), however two auxiliary images were used to train Multi-ABN and generate the fetching instruction.
Qualitative results of our method are illustrated in Fig. \ref{fig:wrs_samples}. The Multi-ABN can be applied in a framework (see subfigure (a)) where the generated sentence are instructed to a robot. While completing the fetching task, the robot collect additional data that are used to generate more accurate or new fetching instructions.
Subfigures (b) and (c) present correct fetching instructions, while  subfigure (d) show an erroneous sentence generation. Indeed in the latter, the referring expression does not allow to disambiguate the target from the other large bottle.

\subsubsection{Visualization of attention maps}
\begin{figure*}[h]
 \subfloat[Bring me the small item on the right-sided armchair.]{\includegraphics[scale = 0.34]{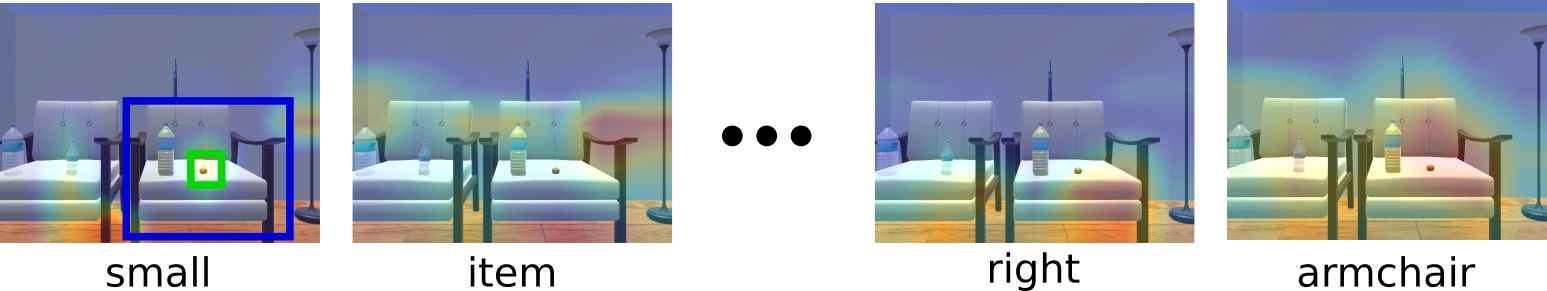}}\\
 \subfloat[Pick up the yellow toy from the white shelf.]{\includegraphics[scale = 0.34]{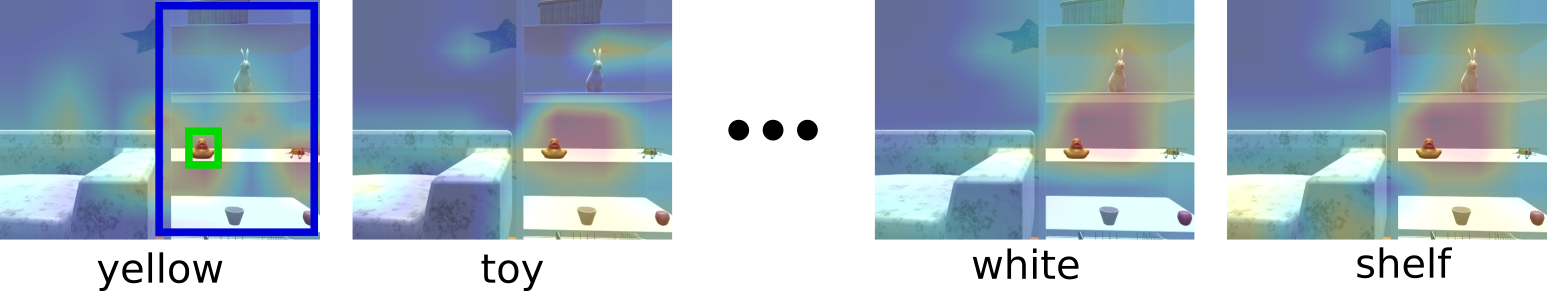}}\\
 \subfloat[Take the tea on the lower row of the shelf.]{\includegraphics[scale = 0.34]{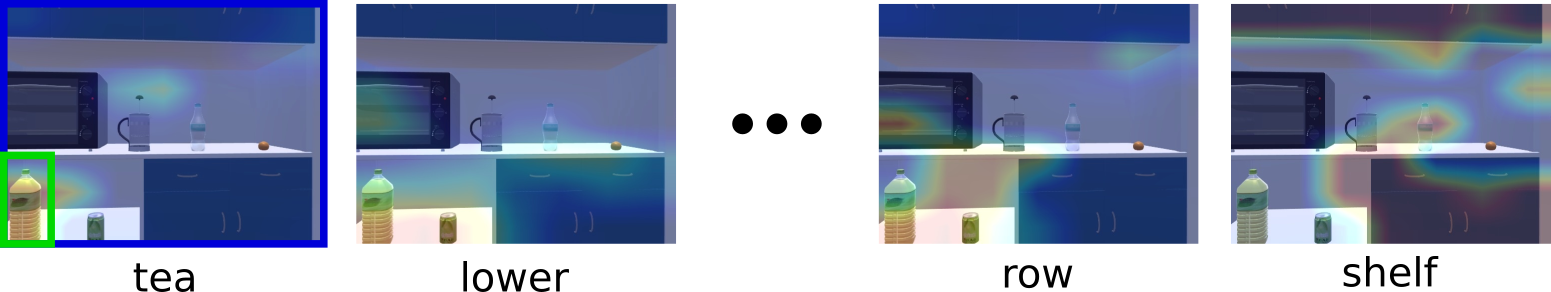}}\\
\caption{\small Multi-ABN visual attention evolution for different sentence steps of sentence generation. The visual attention is updated to the relevant parts of the image. }
    \label{fig:aligment}
\end{figure*}

Similarly to methods based on visual attention, Multi-ABN can also exhibit the visual alignment between text and image. These alignments are depicted in Fig. \ref{fig:aligment}, where the attention map for each generated word is depicted on the image ${\bf x}_v^{1}$.

These results confirm the relevance of the visual attention for generating a sequence of words. Multi-ABN is able to learn the correspondence between linguistic and visual features. Indeed the visual attention was able to focus  on the relevant part of the image. 

\section{Conclusion}
\label{sec:conclusion}
Motivated by the development of communicative DSRs, we developed a method for generating natural fetching instructions. The proposed method (Multi-ABN) is a multimodal attention branch network that generates natural sentences including referring expressions. We summarize the following important contributions of the paper:

\begin{itemize}
 \item The Multi-ABN is an attention branch network that generates sentences based on visual and linguistic attention. The Multi-ABN yields better under BLEU, ROUGE and CIDEr metrics than a baseline method such as \cite{yu2017joint}.  \item The Multi-ABN outperforms visual only  or linguistic only attention branch networks, which emphasizes the contribution of both linguistic and visual modalities.
\item The Multi-ABN is able to generate perspective-free fetching instructions by the use of several visual attention branches related to different viewpoints of the same scene. 
\end{itemize}

In future work, we plan to extend our work with a physical experimental study with non-expert users.   Additionally, we plan to apply the Multi-ABN on a fully communicative DSR scheme that couples sentence generation and natural language comprehension  for fetching tasks.

\clearpage
\acknowledgments{This work was partially supported by JST CREST and SCOPE.}


\bibliography{example}  

\end{document}